\documentclass{article}

\PassOptionsToPackage{numbers, compress}{natbib}

\usepackage[final]{neurips_2025}

\usepackage[utf8]{inputenc} 
\usepackage[T1]{fontenc}    
\usepackage{hyperref}       
\usepackage{url}            
\usepackage{booktabs}       
\usepackage{amsfonts}       
\usepackage{nicefrac}       
\usepackage{microtype}      
\usepackage{xcolor}         
\usepackage{amsmath}
\usepackage{graphicx}
\usepackage{subcaption}
\usepackage{multirow}
\usepackage{array}
\usepackage{xcolor}
\usepackage{adjustbox} 
\usepackage{tikz} 
\usetikzlibrary{calc} 
\usepackage{booktabs}  
\usepackage{cite}  
\usepackage{natbib}  
\usepackage{xpatch}
\usepackage{algorithm}
\usepackage{algpseudocode}
\usepackage{enumitem}

\usepackage{xpatch}
\makeatletter
\xapptocmd{\NAT@bibsetnum}{\setlength{\leftmargin}{0pt}\setlength{\itemindent}{\labelwidth}\addtolength{\itemindent}{\labelsep}}{}{}
\makeatother

\title{Seemingly Redundant Modules Enhance Robust Odor Learning in Fruit Flies}

\author{%
  Haiyang Li\textsuperscript{1*} \hspace{0.3cm}
  Liao Yu\textsuperscript{2*}\hspace{0.3cm}
  Qiang Yu\textsuperscript{3\textdagger} \hspace{0.3cm}
  Yunliang Zang\textsuperscript{1,4\textdagger}\\
  \textsuperscript{1}Academy of Medical Engineering and Translational Medicine, Tianjin University, Tianjin, China \\
  \textsuperscript{2}School of Mathematical Sciences, Beihang University, Beijing, China\\
  \textsuperscript{3}School of Artificial Intelligence, Tianjin University, Tianjin, China \\
  \textsuperscript{4}Xiamen Intretech Inc, Xiamen, Fujian, China \\
  \texttt{haiyangli@tju.edu.cn}\\
  \texttt{yuliao\_16@buaa.edu.cn}\\
  \texttt{yuqiang@tju.edu.cn}  \\
  \texttt{yunliangzang@tju.edu.cn} 
}

\begin{document}

\maketitle
\begingroup
\renewcommand\thefootnote{\textsuperscript{*}} 
\footnotetext{Equal contribution.}
\renewcommand\thefootnote{\textsuperscript{\textdagger}} 
\footnotetext{Corresponding author.}
\endgroup

\begin{abstract}
Biological circuits have evolved to incorporate multiple modules that perform similar functions. In the fly olfactory circuit, both lateral inhibition (LI) and neuronal spike frequency adaptation (SFA) are thought to enhance pattern separation for odor learning. However, it remains unclear whether these mechanisms play redundant or distinct roles in this process. In this study, we present a computational model of the fly olfactory circuit to investigate odor discrimination under varying noise conditions that simulate complex environments. Our results show that LI primarily enhances odor discrimination in low‑ and medium‑noise scenarios, but this benefit diminishes and may reverse under higher‑noise conditions. In contrast, SFA consistently improves discrimination across all noise levels. LI is preferentially engaged in low‑ and medium‑noise environments, whereas SFA dominates in high‑noise settings. When combined, these two sparsification mechanisms enable optimal discrimination performance.  This work demonstrates that seemingly redundant modules in biological circuits can, in fact, be essential for achieving optimal learning in complex contexts.   The code is available at: \url{https://github.com/L-0cean/Fly-SNN}.
\end{abstract}

\section{Introduction}

Biological redundancy is commonly observed in the brain, where different regions, pathways, or mechanisms can perform similar functions \citep{mysore2012reciprocal,driscoll2024flexible,hulse2021connectome,ZANG20231818}. Different motifs of biological redundancy may exist; for instance, different modules may have evolved to fulfill similar roles, ensuring robust neural functions under pathological conditions \citep{cardin2019functional,kroes2012dynamic,jones2017motor,nishimura2012cortical}. Alternatively, these mechanisms may be required to achieve near‑optimal learning in more complex environments. Comparative analyses of the roles of putatively redundant modules in learning can clarify how the brain adapts and, in turn, inform theories that guide neuromorphic design \citep{ganguly2024spike,yik2025neurobench}. For instance, core computational features of the fly olfactory circuit have motivated the FlyLoRA architecture \citep{zou2025flyloraboostingtaskdecoupling}, which enhances task decoupling and parameter efficiency.

The fly olfactory system, a canonical cerebellum‑like circuit, is a tractable model for dissecting neural computation owing to its relative simplicity and well‑characterized anatomy and function \citep{kee2015feed,keene2007drosophila,couto2005molecular,zheng2022structured,manoim2025nonlinear,assisi2020optimality,ZANG2023102765}. Two motifs—lateral inhibition (LI) and spike‑frequency adaptation (SFA)—are prominent in the mushroom body and related circuits and both are known to shape neuronal responses \citep{wilson2013early,busto2010olfactory,nagel2011biophysical}. LI is mediated by inhibitory interneurons that constrain the spatial spread of excitation and sharpen population representations \citep{franco2017reduced,large2016balanced}. SFA reflects spike‑triggered neuronal adaptation currents that accumulate during sustained stimulation, producing a progressive reduction in firing rate and emphasizing stimulus onsets and changes \citep{farkhooi2013cellular,koch2025spike}.

Prior work has extensively characterized the roles of LI and SFA in neural information processing. LI drives competition and winner‑take‑all dynamics \citep{stopfer2005olfactory}, increases population sparseness \citep{wilson2005role}, enhances input contrast \citep{del2025lateral}, and decorrelates activity patterns \citep{giridhar2011timescale}, it also facilitates subtractive or divisive gain modulation and strengthens regularization \citep{olsen2008lateral}. SFA functions as a nonlinear high‑pass filter \citep{benda2021neural}, promoting intensity‑invariant coding \citep{prescott2008spike}, encoding changes in input statistics \citep{clemens2018fast}, sparsifying temporal responses, and forming short-term memory with non-stored retrieval (distinct from synaptic memory) \citep{koch2025spike}.

Although the roles of LI and SFA in sparsifying neuronal responses are well documented, their relative contributions to learning—particularly under naturalistic conditions—remain less well characterized. Both mechanisms are expected to transform odor inputs into spatially and temporally sparse codes, attenuate noise, and enhance pattern separation \citep{barnum2022olfactory,gregor2011structured,druckmann2012mechanistic,luo2010generating,chalk2018toward}(Figure $\ref{Figure 1}$). However, how they shape noisy odor representations and support discrimination learning across different noise regimes is not well understood. From an AI perspective, designing noise‑robust classifiers is a classical challenge; elucidating how a fly learns to classify noisy odors may inform the development of robust, biologically inspired algorithms \citep{lamy2019noise}.

In this work, we developed a fly olfactory circuit model to investigate the roles of LI and SFA in odor discrimination tasks under varying noise conditions that simulate complex environments. Our main finding is that LI enhances learning performance in low- and medium-noise conditions, but this benefit gradually diminishes and may reverse when odors become noisier. In contrast, SFA consistently improves odor learning regardless of noise levels. LI tends to be more effective in low‑ and medium‑noise environments, while SFA shows superior performance under high‑noise conditions. When combined, the enhancement effects of these two mechanisms can be added up to achieve the optimal performance. These results suggest that seemingly redundant modules may be selectively recruited to optimize learning under different noisy conditions.

\section{Method Overview}
We developed a spiking neural network model of the fly olfactory circuit. The input layer consists of 50 olfactory receptor neurons (ORNs) that transduce odor stimuli into spikes. Each ORN projects to a corresponding projection neuron (PN), yielding 50 PNs. We also included local interneurons (LNs), which receive excitatory input from ORNs and provide lateral inhibition onto PNs. PN activity is relayed to 2,000 Kenyon cells (KCs) in the mushroom body; each KC samples inputs from approximately six PNs on average. Finally, KCs converge onto mushroom body output neurons (MBONs), which serve as readout units; in our task configuration, each MBON corresponds to an odor class being learned \citep{ng2002transmission,masse2009olfactory}.

The fly olfactory pathway exhibits three canonical features that are critical for discrimination \citep{zou2025structural}: large expansion ($\text{PN} \to \text{KC}$), sparse connectivity, and sparse coding. Here, we focus on the role of spatiotemporally sparse spike coding, which conventional ANN models cannot capture; other architectural and connectivity parameters were constrained to experimentally observed values. 

\begin{figure}
  \centering
  \includegraphics[width=0.8\textwidth]{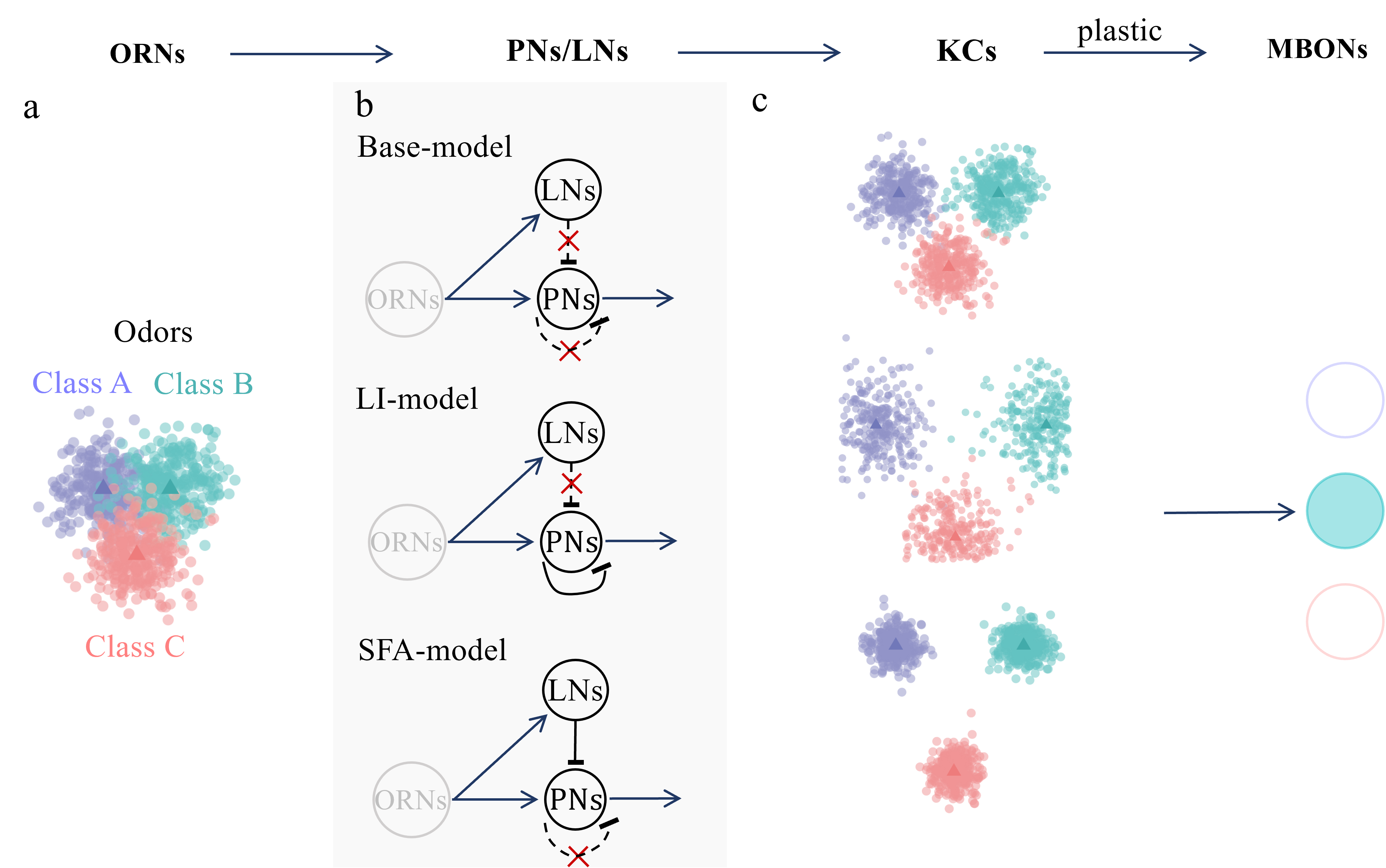}
  \caption{ \textbf{Schematic of the fly olfactory circuit model.} Odor inputs are sensed and encoded by ORNs. After passing through the ORNs, odor-triggered spikes in PNs can be shaped by two factors: LI caused by LNs and SFA by inherent adaptation currents. Neuronal spikes in KCs may subsequently show different variation patterns to favor odor discrimination in MBONs.}
  \label{Figure 1}
\end{figure}

A schematic of the network and its putative role in odor discrimination is shown in Figure $ \ref{Figure 1} $. After the KC stage, the sparsification mechanisms can transform odor responses in ways that facilitate odor discrimination. For example, they may preserve the original inter‑class separability while increasing intra‑class compactness (top), increase inter‑class separation while keeping intra‑class compactness unchanged (middle), or simultaneously achieve high intra‑class compactness and improved inter‑class separability (bottom). Regardless of the specific transformation, these changes are supposed to enhance the decision boundaries for odor discrimination. 

Synaptic plasticity was restricted to the KC→MBON connections; all other synapses were fixed. Details of the training procedure and plasticity rule are provided in the \nameref{sec:learning_algorithm} section.

\paragraph{Input Data.}
We adapted the odor-discrimination dataset from \citep{wang2021evolving}. For each odor class $i$, we define a prototype ORN response vector $\mathbf{I}_i \in \mathbb{R}^{50}$. Its components are independently sampled from a uniform distribution: $I_{i,j} \sim \text{U}(0, 1)$ for $j = 1, \dots, 50$. An individual sample from class $i$, denoted $\mathbf{I}_i^{(k)}$, is generated by adding an additive noise vector to the prototype: $\mathbf{I}_i^{(k)} = \mathbf{I}_i + \mathbf{Y}^{(k)}$, where every component of the noise vector $\mathbf{Y}^{(k)} \in \mathbb{R}^{50}$ is independently sampled from a zero-mean Gaussian distribution, $Y^{(k)}_j \sim \mathcal{N}(0, \sigma^2_{\text{noise}})$. In accordance with the non-negativity of ORN firing rates, we applied element-wise clipping so that $\mathbf{I}_i^{(k)} \ge 0$.

\paragraph{Neuron Model.}
All neurons were modeled as leaky integrate-and-fire (LIF) units with a soft reset. The membrane potential of neuron $j$ in the layer $X$ evolves according to:
\begin{equation}
\tau_{\mathrm{m}}^X \frac{\mathrm{d}V_j^X(t)}{\mathrm{d}t} = -V_j^X(t) + I_{\text{input},j}^X(t) + I_{\text{bias},j}^X + I_{\text{SFA},j}^X(t) + I_{\text{LI},j}^X(t)
\end{equation}
Where $\tau_{\mathrm{m}}^X$ is the membrane time constant. $I_{\text{input},j}^X(t)$ denotes the input current to neuron $j$ in layer $X$, arising from odor-evoked stimulation or synaptic drive from presynaptic neurons; $I_{\text{bias},j}^X$ is a constant bias that sets a baseline drive (used primarily in PNs and LNs to provide a small excitatory input); $I_{\text{SFA},j}^X(t)$ is a spike-triggered adaptation current that depends on the neuron’s recent spiking history and typically exerts a net inhibitory effect; and $I_{\text{LI},j}^X(t)$ is the lateral inhibitory current from LNs onto PNs, scaled by LN spiking activity.

In the model, a spike is emitted when $V_j^X(t)$ reaches the threshold $V_{\text{th}}^X$. Upon spiking, the membrane potential undergoes a soft reset: $V_j^X(t^+) = V_j^X(t) - V_{\text{th}}^X$. For simplicity, we used the same membrane time constant and threshold across layers ($\tau_\mathrm{m}^X = \tau_\mathrm{m}, V_{\text{th}}^X = V_{\text{th}}$), except in the readout (MBON) layer, where the threshold was set to $1.5$ times higher. Given the dense $\text{KC} \to \text{MBON}$ connectivity, this higher threshold mitigates excessive spiking driven by the convergence of inputs from thousands of KCs.

\paragraph{LI Mechanism.}
LI onto PNs is mediated by LNs and modeled as an inhibitory current driven by an exponentially decaying trace of LN spiking. The inhibitory current onto PN $j$ is:
\begin{equation}
\label{eq:LI}
I_{\text{LI},j}^{\text{PN}}(t) = \sum_{k} w_{\text{LN}\to\text{PN},jk} T_k^{\text{LN}}(t)
\end{equation}
Where $w_{\text{LN}\to\text{PN},jk} \le 0$ are inhibitory synaptic weights, and $T_k^{\text{LN}}(t)$ is the spike-triggered trace of LN $k$ that integrates recent activity and decays over time:
\begin{align}
\tau_{\text{trace\_LN}}\frac{\mathrm{d}T_k^{\text{LN}}(t)}{\mathrm{d}t} &= -T_k^{\text{LN}}(t) + S_k^{\text{LN}}(t) \\
S_k^{\text{LN}}(t) &= \sum_{l} \delta(t - t_{kl}^{\text{LN}})
\end{align}
with $\tau_{\text{trace\_LN}}$ the decay time constant, $S_k^{\text{LN}}(t)$ the spike train of LN $k$, and $t_{kl}^{\text{LN}}$ the $l$-th spike time. To compensate for the added inhibition and maintain comparable PN firing rates across conditions, we apply a modest increase in PN input drive when LI is enabled.

\paragraph{SFA Mechanism.}
SFA was implemented in PNs, LNs, and KCs as an inhibitory, spike-triggered current driven by an exponentially decaying state variable. For neuron $j$ in layer $X$,
\begin{align}
\label{eq:SFA}
I_{\text{SFA},j}^X(t) &= w_{\text{SFA}}^X A_j^X(t) \\
\tau_{\text{SFA}}^X \frac{\mathrm{d}A_j^X(t)}{\mathrm{d}t} &= -A_j^X(t) + S_j^X(t) \\
S_j^X(t) &= \sum_{l} \delta(t - t_{j,l}^X)
\end{align}
where $A_j^X(t)$ integrates the recent spiking history and decays with time constant $\tau_{\text{SFA}}^X$; $w_{\text{SFA}}^X \le 0$ sets the strength (sign) of the inhibitory adaptation current; $S_j^X(t)$ is the spike train of neuron $j$ (Dirac delta representation), and $t_{j,l}^X$ denotes the $l$-th spike time.

To maintain comparable firing rates across conditions when SFA is enabled, we applied a small positive bias current to PNs and LNs.

\paragraph{Learning Algorithm.} \label{sec:learning_algorithm}

In our model, classification is based on the average membrane potential of the MBONs over a fixed evaluation window rather than on spike counts. Membrane potentials vary more smoothly than discrete spike trains and thus provide a more stable signal for gradient-based optimization \citep{guo2022loss,shen2024rethinking}. The complete training procedure is outlined in Algorithm \ref{alg:mylabel}.

\begin{algorithm}
\caption{Simplified SNN Training with Adaptive Mechanisms for Olfactory Tasks}
\label{alg:mylabel}
\begin{algorithmic}[1]
\State \textbf{Input:} Training dataset $D_{\text{train}}$, Learnable weights $W_{\text{KC}\to\text{MBON}}$, Non-learnable parameters $P$.

\For{epoch = 1, 2, ..., num\_epochs}
  \For{each batch $(x_{\text{batch}}, y_{\text{batch}})$ in $D_{\text{train}}$}
    
    \State \textbf{Phase 1: Temporal Forward Simulation}
    \State Initialize all SNN states $H_0$. Set $H_{\text{hist}} = [ ]$.
    \For{step $t=0$ to num\_steps - 1}
      \State $I_t = \operatorname{Input}(x_{\text{batch}}, \text{step})$ \Comment{Get input current}
      \State $H_{t+1} = F(I_t, H_t, P)$ \Comment{SNN forward pass and state update}
      \State Record MBON membrane potentials in $H_{\text{hist}}$.
    \EndFor
    
    \State \textbf{Phase 2: Backpropagation and Weight Update}
    \State $\hat{Y}_{\text{batch}} = G(H_{\text{hist}})$ \Comment{Process MBON output}
    \State loss = $L(\hat{Y}_{\text{batch}}, y_{\text{batch}})$ \Comment{Cross-Entropy Loss}
    \State loss.backward() \Comment{Perform Backpropagation Through Time}
    \State Optimizer.step() \Comment{Update $W_{\text{KC}\to\text{MBON}}$}
    
  \EndFor
\EndFor

\end{algorithmic}
\end{algorithm}

For input sample $i$, we recorded each MBON's membrane potential over the evaluation window $[t_s, t_e]$ and computed its time average. Let $N_{\text{MBON}}$ be the number of MBONs. The mean potential for MBON $j$ and the corresponding vector across MBONs are
\begin{equation}
\bar{V}_{\mathrm{m},i}^j = \frac{1}{t_e - t_s} \int_{t_s}^{t_e} V_j(t)\,\mathrm{d}t, \quad \mathbf{\bar{V}}_{\mathrm{m},i} \in \mathbb{R}^{N_{\text{MBON}}}
\end{equation}
We convert $\mathbf{\bar{V}}_{\mathrm{m},i}$ to class probabilities using a $\operatorname{softmax}$ function and train the network with a cross-entropy loss:
\begin{align}
p_i^j &= \operatorname{softmax}(\mathbf{\bar{V}}_{\mathrm{m},i})_j \\
L_i &= - \sum_{j}^{N_{\text{MBON}}} y_i^j \log(p_i^j)
\end{align}
Where $y_i^j$ is the one-hot target for sample $i$.

We optimized the learnable $\text{KC} \to \text{MBON}$ synaptic weights $W_{\text{KC} \to \text{MBON}}$ using backpropagation through time (BPTT), unrolling the network over the evaluation window. Because LIF spikes arise from a step nonlinearity with zero derivative almost everywhere, we used a surrogate gradient during the backward pass. Specifically, we replaced the derivative of the Heaviside with a smooth arctan-based surrogate:
\begin{equation}
\sigma'(u) \approx \frac{k_1}{1 + (k_2 u)^2}
\end{equation}
where $k_1, k_2 > 0$ are scaling constants.

Training uses Adam with L2 weight decay (weight regularization) to improve generalization. We also employed a $\text{ReduceLROnPlateau}$ scheduler that monitors classification accuracy on a held-out set and reduced the learning rate by a factor $\alpha=0.2$ if accuracy did not improve for $n=10$ epochs.

Each model was trained for $100$ epochs using mini-batches of size $256$. After each batch, the loss was computed, gradients were backpropagated through time, and parameters were updated using the optimizer.

\section{Experiments} \label{sec:experiments}

\paragraph{Dataset and models:} 
We generated an odor dataset comprising 30,000 training samples and 10,000 test samples, each drawn at random as described above. For most simulations, we evaluated three network configurations: (i) the Baseline model, which includes neither LI nor SFA; (ii) the LI model, which adds LI onto PNs; and (iii) the SFA model, which implements SFA in PNs, LNs, and KCs. In Figure $\ref{fig5}$, we also simulated the full model with both LI and SFA.

\paragraph{Experimental settings:} 
Simulations were implemented in Python (snnTorch) and run on an NVIDIA A800 GPU. Network dynamics were simulated with a $1$-ms time step; each trial comprised a $10$-ms pre-stimulus baseline followed by a $30$-ms odor presentation. All neurons shared a membrane time constant of $10$ ms. Firing thresholds were $0.8$ for PNs, LNs, and KCs, and $1.2$ for MBONs. Connectivity featured sparse $\text{PN} \to \text{KC}$ projections (each $\text{KC}$ received inputs from $6$ PNs; fixed weight $0.3$) and fully connected $\text{KC} \to \text{MBON}$ synapses initialized uniformly in $[0, 0.08]$. $\text{LI}$ used a $5$-ms $\text{LN}$ trace time constant, and $\text{SFA}$ used a $50$-ms adaptation time constant. Models were trained for $100$ epochs (batch size $256$) with Adam (initial learning rate $1.0\times 10^{-4}$). Performance was evaluated from the time-averaged MBON membrane potentials over the stimulus window.

\section{Results}
We evaluated our fly olfactory network model on a noisy odor discrimination task by incorporating LI and SFA mechanisms. The results show that these two seemingly redundant mechanisms play complementary roles in learning across varying noise conditions, thereby optimizing odor discrimination.

\subsection{Odor discrimination in noise-free conditions} \label{sec:odor_discrimination_noise_free}

\begin{figure}[htbp]
    \centering
    \begin{tikzpicture}
        \node[anchor=south west,inner sep=0] (image) at (0,0) 
              {\includegraphics[width=0.7\textwidth]{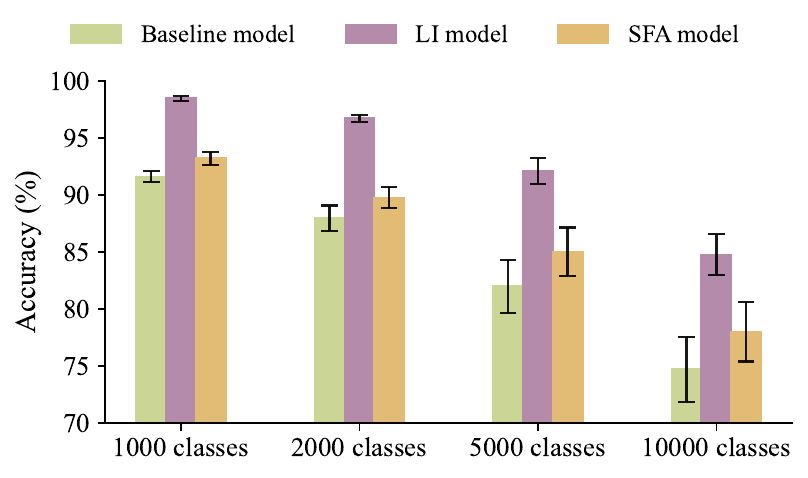}};
    \end{tikzpicture}
    \caption{ \textbf{Odor discrimination accuracy for fly olfactory circuit variants in a noise‑free setting.} Performance is shown for three configurations—--Baseline, LI, and SFA models—--as a function of the number of odor classes (1,000–10,000).}
    \label{Figure 2}
\end{figure}

Our results show that the fly olfactory circuit model learns odor discrimination effectively (Figure \ref{Figure 2}). We first examined discrimination in noise-free conditions. In the baseline model---without LI or SFA---the discrimination accuracy reaches $91.7\%$ for $1,000$ odor classes. Although accuracy decreases as the number of classes increases, it remains $74.72\%$ for $10,000$ odor classes. These findings indicate that the circuit's other features confer strong pattern-classification capacity even without explicit sparsifying mechanisms in the circuit \citep{zou2025structural,zang2023sodium}.

Both LI and SFA improve odor discrimination relative to the Baseline model across all tested numbers of classes. By comparison, the LI model achieves significantly higher accuracy than the SFA model under the same conditions.

\subsection{Odor discrimination in noisy conditions} \label{sec:odor_discrimination_noisy}

Table \ref{tab1} presents a comparative analysis of discrimination performance under varying noise intensities for the Baseline, LI, and SFA models, with odor classes ranging from $1,000$ to $5,000$. At low- and medium-noise levels---defined as noise intensity (N.I.) $< 0.20$ for $1,000$- and $2,000$-class odor discrimination, and N.I. $< 0.15$ for $5,000$-class odor discrimination---the results are consistent with the noise-free context ($\text{LI}$ model $> \text{SFA}$ model $> \text{Baseline}$ model), although overall test accuracy decreases. These findings indicate that sparsification of neuronal spikes in KCs via $\text{LI}$ is more effective than $\text{SFA}$ in enhancing odor discrimination under no- and low-noise conditions. However, when odor inputs become noisier ($\text{N.I.} \ge 0.20$ for $1,000$- and $2,000$-class discrimination, and $\text{N.I.} \ge 0.15$ for $5,000$-class discrimination), the $\text{SFA}$ model consistently achieves the highest discrimination accuracy.

\begin{table}[htbp]
  \centering
  \caption{\textbf{Comparative performance of the Baseline, SFA, and LI models for 1,000-, 2,000-, and 5,000-class odor discrimination under different noise ntensities.} Magenta values indicate the highest performance under each condition, violet values represent the second-best performance, and blue values denote the lowest performance.}
    \begin{tabular}{l|l|ccccccccc}
    \toprule
    \multicolumn{1}{c|}{\multirow{12}{*}{\shortstack{Acc.\\ (\%)}}} & \multicolumn{1}{c|}{\multirow{4}{*}{N.I.}} & \multicolumn{3}{c}{1000 classes} & \multicolumn{3}{c}{2000 classes} & \multicolumn{3}{c}{5000 classes} \\
    \cmidrule(lr){3-5} \cmidrule(lr){6-8} \cmidrule(lr){9-11}
    \multicolumn{1}{c|}{} & & Baseline & LI & SFA & Baseline & LI & SFA & Baseline & LI & SFA \\
    \midrule
    & 0 & \textcolor{blue}{91.70} & \textcolor{magenta}{\textbf{98.30}} & \textcolor{violet}{93.50} & \textcolor{blue}{88.85} & \textcolor{magenta}{\textbf{96.85}} & \textcolor{violet}{90.90} & \textcolor{blue}{82.00} & \textcolor{magenta}{\textbf{92.15}} & \textcolor{violet}{85.06} \\
    & 0.05 & \textcolor{blue}{78.46} & \textcolor{magenta}{\textbf{96.35}} & \textcolor{violet}{82.78} & \textcolor{blue}{68.62} & \textcolor{magenta}{\textbf{92.47}} & \textcolor{violet}{74.81} & \textcolor{blue}{52.22} & \textcolor{magenta}{\textbf{77.47}} & \textcolor{violet}{58.93} \\
    & 0.1 & \textcolor{blue}{74.61} & \textcolor{magenta}{\textbf{91.85}} & \textcolor{violet}{78.26} & \textcolor{blue}{61.70} & \textcolor{magenta}{\textbf{83.07}} & \textcolor{violet}{67.87} & \textcolor{blue}{38.65} & \textcolor{magenta}{\textbf{57.83}} & \textcolor{violet}{45.18} \\
    & 0.15 & \textcolor{blue}{74.04} & \textcolor{magenta}{\textbf{83.63}} & \textcolor{violet}{79.94} & \textcolor{blue}{61.14} & \textcolor{magenta}{\textbf{71.52}} & \textcolor{violet}{68.54} & \textcolor{blue}{34.47} & \textcolor{violet}{42.15} & \textcolor{magenta}{\textbf{43.89}} \\
    & 0.2 & \textcolor{blue}{72.64} & \textcolor{violet}{74.78} & \textcolor{magenta}{\textbf{78.77}} & \textcolor{blue}{58.80} & \textcolor{violet}{58.87} & \textcolor{magenta}{\textbf{67.34}} & \textcolor{violet}{32.70} & \textcolor{blue}{31.91} & \textcolor{magenta}{\textbf{43.04}} \\
    & 0.25 & \textcolor{violet}{67.32} & \textcolor{blue}{62.63} & \textcolor{magenta}{\textbf{74.34}} & \textcolor{violet}{52.26} & \textcolor{blue}{45.84} & \textcolor{magenta}{\textbf{61.55}} & \textcolor{violet}{27.74} & \textcolor{blue}{22.26} & \textcolor{magenta}{\textbf{36.94}} \\
    & 0.3 & \textcolor{violet}{59.03} & \textcolor{blue}{53.82} & \textcolor{magenta}{\textbf{69.34}} & \textcolor{violet}{43.34} & \textcolor{blue}{37.50} & \textcolor{magenta}{\textbf{55.78}} & \textcolor{violet}{20.26} & \textcolor{blue}{16.55} & \textcolor{magenta}{\textbf{30.38}} \\
    \bottomrule
    \end{tabular}
  \label{tab1}
\end{table}

\begin{figure}[htbp]
    \centering
    \begin{tikzpicture}
        \node[anchor=south west,inner sep=0] (image) at (0,0) {\includegraphics[width=1\textwidth]{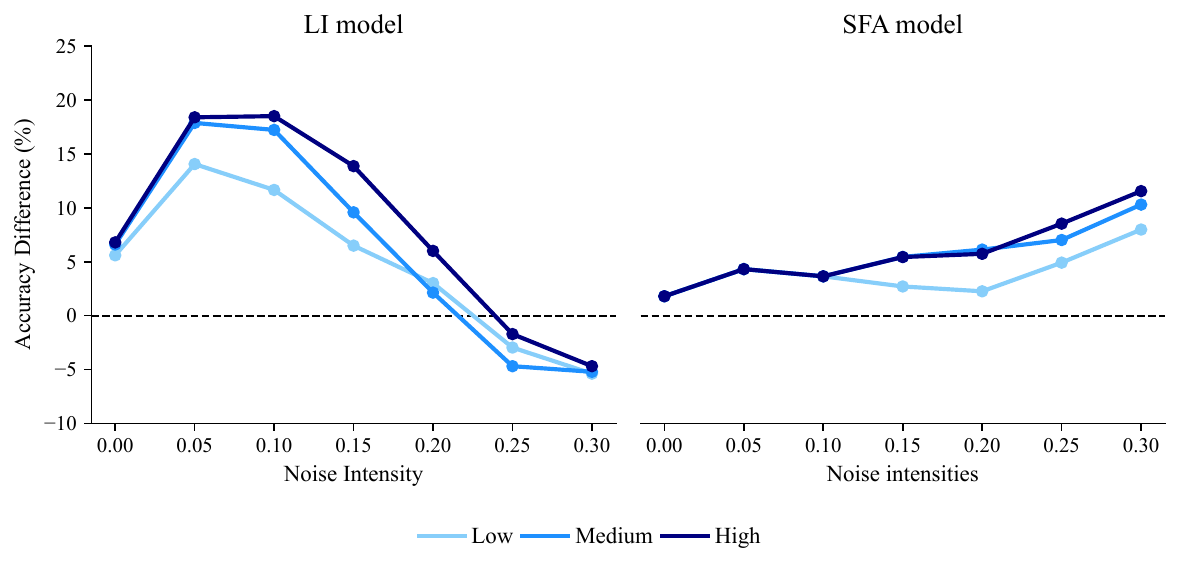}}; 

    \end{tikzpicture}

    \caption{\textbf{Changes in discrimination performance of the SFA and LI models relative to the Baseline model across varying degrees of inhibition and adaptation under different noise intensities.} Results are shown for 1,000-class odor discrimination only. The ``Low,'' ``Medium,'' and ``High'' conditions (distinguished by color) represent increasing strengths of the respective mechanisms, achieved by systematically adjusting the relevant synaptic weights: $w_{\text{LN} \to \text{PN}}$ in Eq.\ \ref{eq:LI} for LI, and $w_{\text{SFA,X}}$ in Eq.\ \ref{eq:SFA} for SFA. For both mechanisms, ``Low,'' ``Medium,'' and ``High'' correspond to weight increases in an approximate 1:2:3 ratio.}
    \label{fig4}
\end{figure}

We systematically analyzed the impact of the strength of each sparsification mechanism---LI and SFA---on discrimination performance (Figure \ref{fig4}). The simulation results indicate that both facilitation effects depend on noise levels, but in distinct ways. For LI, discrimination accuracy initially improves as \text{N.I.} increases, but then gradually diminishes and can even reverse at higher noise levels. Across most of the tested \text{N.I.} range, stronger inhibition promotes discrimination accuracy. Compared to the Baseline model, strong inhibition improves accuracy by 6.80\% when $\text{N.I.} = 0.0$ and 18.52\% when $\text{N.I.} = 0.1$. Notably, although LI impairs discrimination at high \text{N.I.},  ($-4.5\%$ when $\text{N.I.} = 0.30$), the reduction is still less pronounced for stronger inhibition.

In contrast to LI, SFA consistently facilitates odor discrimination across all tested \text{N.I.} ranges. As with LI, greater SFA strength generally yields higher accuracy, with improvements of 1.80\% when $\text{N.I.} = 0.0$, 3.65\% when $\text{N.I.} = 0.1$, and 11.56\% when $\text{N.I.} = 0.30$.

\begin{figure}[htbp]
    \centering
    \includegraphics[width=1\textwidth]{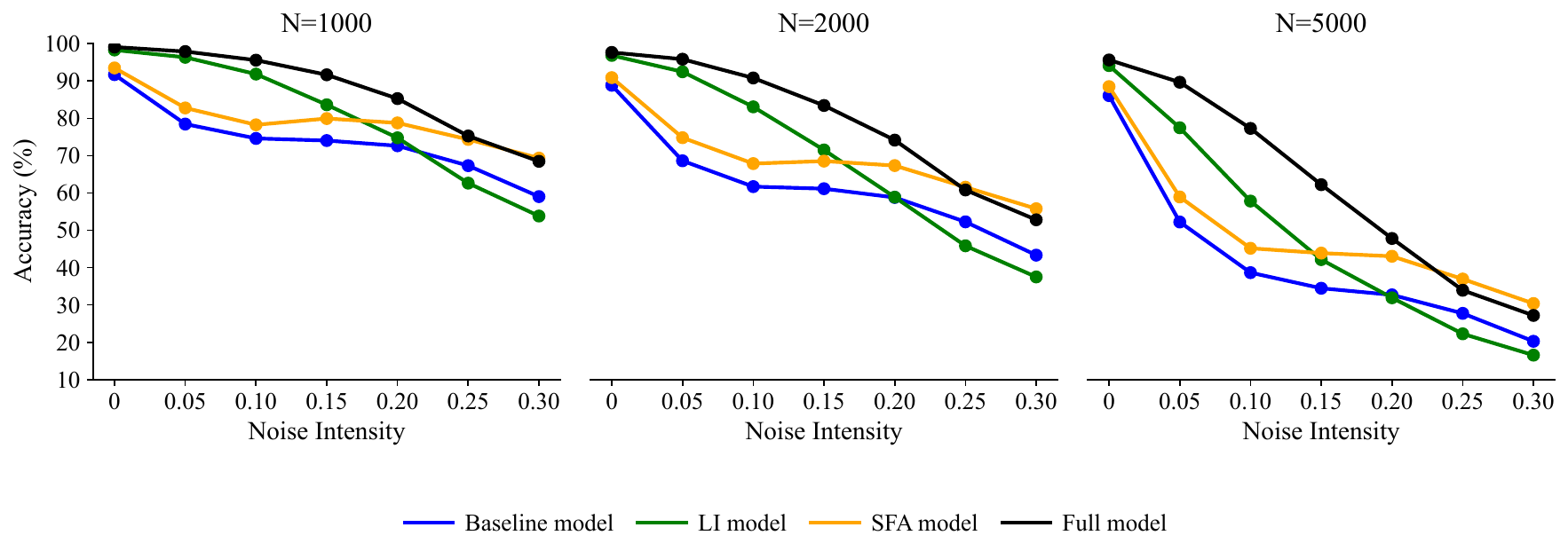} 

    \caption{\textbf{Discrimination performance of the SFA model, LI model, Full (SFA + LI) model, and Baseline model under different noise intensities.} The models were tested on odor discrimination tasks with 1,000, 2,000, and 5,000 odor classes.}
    \label{fig5} 
\end{figure}

In addition, we investigated whether LI and SFA could be dynamically combined to achieve optimal odor discrimination learning (Figure \ref{fig5}). Our results show that orchestrating these two mechanisms produces higher discrimination accuracy than using either mechanism alone under low- and medium-noise levels. Only under high-noise conditions does the SFA model outperform all other models. The benefits of these two mechanisms are additive. These findings suggest that, although each mechanism is most effective within specific noise regimes, LI and SFA can work together in a complementary manner, combining their individual effects to optimize the learning process.

\subsection{Learning speed}

\begin{figure}[htbp]
    \centering
    \includegraphics[width=1\textwidth]{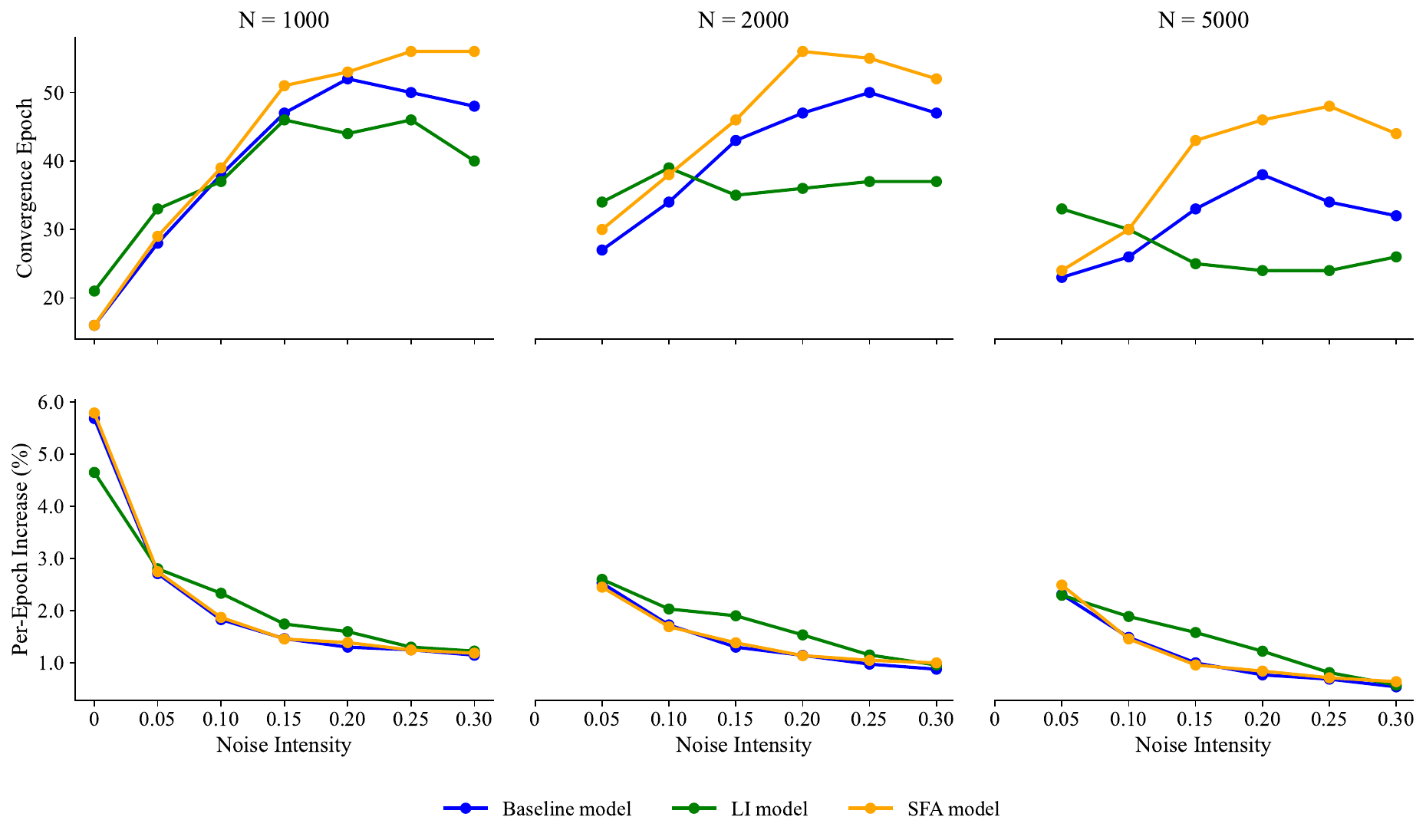} 

    \caption{\textbf{Impacts of LI and SFA on model convergence speed.} The top (bottom) panel shows the number of training epochs required for convergence (the accuracy gain per epoch) for the Baseline, LI, and SFA models, plotted against noise intensity for different odor category sizes: 1,000, 2,000, and 5,000 classes. To avoid potential misinterpretations from relying solely on maximum accuracy and to objectively assess learning progress, convergence is defined as the point at which model accuracy growth plateaus. Specifically, a model is considered converged when the average improvement in accuracy over $ n = 10 $ consecutive epochs falls below a predefined threshold (threshold = $0.003$). }
    \label{fig10} 
\end{figure}

To further assess the impacts of LI and SFA on learning efficiency, we provide a quantitative analysis of how these two sparsification mechanisms shape the time course of odor discrimination learning. As shown in Figure \ref{fig10} (top), similar to the final discrimination accuracy values (Table $\ref{tab1}$), the number of training epochs required for convergence depends on noise level. In the low‑noise range, the LI model requires more epochs to reach the defined converged state, regardless of odor category size. SFA follows LI, while the Baseline model reaches convergence fastest. The convergence speed can be explained by combining the final discrimination accuracy (Table $\ref{tab1}$) with the accuracy gain per epoch (Figure \ref{fig10}, bottom). In the low-noise range, both LI and SFA require higher final accuracy thresholds to converge, but their per‑epoch gains show no advantage over the Baseline model, resulting in slower convergence.

In higher‑noise ranges, however, the LI model exhibits the fastest convergence, followed by the Baseline model, with the SFA model consistently the slowest. In this regime, the LI model generally maintains an advantage over the Baseline model in per‑epoch accuracy gain, explaining its faster convergence (Figure \ref{fig10}, top). When N.I. = 0.3, the per‑epoch accuracy gain is similar across models; however, SFA requires a higher final accuracy threshold to reach convergence compared to both the Baseline and LI models (Table $\ref{tab1}$). Consequently, LI reaches the converged state first, the Baseline model second, and the SFA model last.  

These results suggest that, depending on the noise level, learning speed—alongside discrimination accuracy—may be an important factor in determining which sparsification mechanism is recruited during noisy odor discrimination learning.

\subsection{Sensitivity analysis}

\subsubsection{Sensitivity to noise types}

The previous results were obtained by simulating the odor discrimination task with Gaussian noise. To assess the generalizability of our findings, we evaluated the effects of the sparsification mechanisms under different noise environments. In addition to Gaussian noise, we simulated noise generated by an Ornstein–Uhlenbeck (OU) process, which captures the temporal correlations often observed in natural odors. Odor samples with OU noise were generated using a procedure analogous to that used for Gaussian noise.  

The results, summarized in Table $\ref{tab4}$, show that the effects of LI and SFA on discrimination performance with OU noise are generally consistent with those observed using Gaussian noise. LI achieves the highest discrimination accuracy at low- and medium-noise levels, whereas SFA is most effective at higher noise intensities. The main difference is that, within the tested noise range, LI—like SFA—consistently improves the discrimination of noisy odors compared to the Baseline model, rather than impairing performance at high noise levels, although its benefit gradually diminishes. Overall, these findings confirm that the complementary effects of LI and SFA are robust across the tested noise types.

\begin{table}[htbp]
\centering
\caption{\textbf{Comparative performance of the Baseline, LI, and SFA models for 1,000-class odor discrimination under different OU noise intensities. }}
\begin{tabular}{lc|ccc} 
\toprule
 & N.I. & Baseline & LI & SFA \\
\midrule
\multirow{5}{*}{\begin{tabular}[c]{@{}l@{}}Acc.\\ (\%)\end{tabular}} 
& 0.1 & \textcolor{blue}{86.02} & \textcolor{magenta}{\textbf{97.98}} & \textcolor{violet}{88.66} \\ 
& 0.3 & \textcolor{blue}{78.66} & \textcolor{magenta}{\textbf{96.43}} & \textcolor{violet}{82.55} \\
& 0.5 & \textcolor{blue}{76.01} & \textcolor{magenta}{\textbf{93.10}} & \textcolor{violet}{78.85} \\
& 0.9 & \textcolor{blue}{74.54} & \textcolor{magenta}{\textbf{87.84}} & \textcolor{violet}{78.50} \\
& 1.5 & \textcolor{blue}{67.55} & \textcolor{violet}{69.72} & \textcolor{magenta}{\textbf{77.49}} \\
\bottomrule
\end{tabular}
\label{tab4} 
\end{table}

\subsubsection{Parameter sensitivity analysis}
To further verify the reliability of our simulation results, we conducted a comprehensive parameter sensitivity analysis. Three key hyperparameters—random seed, learning rate, and batch size—were varied while keeping all other experimental conditions fixed to assess their impacts on odor discrimination.

Our tests reveal that:
\begin{itemize}[label=\Large\textbullet]
    \item Varying the random seed typically changed accuracy by less than 1.3\%.
    \item Adjusting the learning rate to 0.5–2.0 of its default value resulted in accuracy changes generally under 1.6\%.
    \item Scaling the batch size to 0.5–2.0 of its standard value produced accuracy differences typically below 3.0\%.
\end{itemize}

These results demonstrate that, within reasonable variation ranges, model performance is highly robust to hyperparameter choices. Importantly, across all parameter variations, our core conclusions remain unchanged: the LI model achieves optimal performance in low- and medium-noise environments, whereas the SFA model performs best under high-noise conditions. This analysis confirms that our main findings are reliable and independent of specific hyperparameter settings, providing a strong foundation for the interpretations presented in this study.

\section{Conclusion}
In this paper, we present a computational demonstration of how two distinct neural sparsification mechanisms---circuit‑level LI and neuronal‑level SFA---provide complementary advantages for noisy odor discrimination. Both mechanisms, well‑documented in biological neural systems, actively shape and refine spatiotemporal neuronal responses. Our simulation results reveal a noise‑dependent recruitment pattern: LI is preferentially engaged under low‑noise conditions, whereas SFA dominates in high‑noise environments. The fly can orchestrate these two sparsification mechanisms to achieve optimal discrimination performance. These findings illustrate how seemingly redundant circuit modules in biological systems may, in fact, represent an optimized strategy for maintaining robust learning performance across diverse environmental conditions. This study sits at the interface of computational neuroscience and spiking neural networks, leveraging AI methods to investigate learning processes in the fly olfactory circuit. While this interdisciplinary approach is enriching, it also carries several limitations, as discussed below. 

\section{Limitations}
We present a computational model of the fly olfactory circuit and investigate the roles of LI and SFA in odor discrimination learning. While our findings offer valuable insights, several limitations point to promising directions for future research.

First, regarding the training and test datasets, our simulations used artificially generated odors \citep{wang2021evolving}. Although the odor generation method was experimentally inspired and provided a controlled environment, future work should validate odor discrimination performance using more naturalistic and physiologically realistic odor stimuli.  

Second, our current model assumes that plasticity is confined to synaptic connections between KCs and MBONs. The synaptic update rule follows backpropagation through time, which is generally considered biologically implausible. While we believe this does not alter our conclusions, future studies should explore biologically plausible learning algorithms. Moreover, as in most cerebellum‑like circuit models, the synaptic weights between the input and hidden layers (ORN–KC synapses) are fixed \citep{ZANG2023102765,zang2019climbing}. These connections may also undergo other forms of plasticity, such as unsupervised learning via Oja’s rule \citep{oja1982simplified}. Exploring these possibilities would provide a more comprehensive understanding of how learning unfolds across the entire circuit.  

Finally, although our parameter robustness analysis and exploration of different noise types support the generalizability of our findings, the parameter ranges and noise characteristics examined remain limited. Broader investigations into diverse environmental conditions and increased model complexity could further strengthen the conclusions.  

\section{Acknowledgement}
This work was supported by the National Key Research and Development Program of China (2023YFF1204200), the National Natural Science Foundation of China (62476197, 12372060, 92370103, 62176179), and the Xiaomi Foundation. The funders had no role in study design, data collection and analysis, decision to publish, or preparation of the manuscript.

\bibliographystyle{unsrtnat}
\small


\normalsize


\newpage
\section*{NeurIPS Paper Checklist}


\begin{enumerate}

\item {\bf Claims}
    \item[] Question: Do the main claims made in the abstract and introduction accurately reflect the paper's contributions and scope?
    \item[] Answer: \answerYes{} 
    \item[] Justification: We have summarized the main contributions and scope in the abstract and introduction sections.
    \item[] Guidelines:
    \begin{itemize}
        \item The answer NA means that the abstract and introduction do not include the claims made in the paper.
        \item The abstract and/or introduction should clearly state the claims made, including the contributions made in the paper and important assumptions and limitations. A No or NA answer to this question will not be perceived well by the reviewers. 
        \item The claims made should match theoretical and experimental results, and reflect how much the results can be expected to generalize to other settings. 
        \item It is fine to include aspirational goals as motivation as long as it is clear that these goals are not attained by the paper. 
    \end{itemize}

\item {\bf Limitations}
    \item[] Question: Does the paper discuss the limitations of the work performed by the authors?
    \item[] Answer: \answerYes{} 
    \item[] Justification: We mentioned some limitations related to model performance in Appendix $\ref{Performance}$.
    \item[] Guidelines:
    \begin{itemize}
        \item The answer NA means that the paper has no limitation while the answer No means that the paper has limitations, but those are not discussed in the paper. 
        \item The authors are encouraged to create a separate "Limitations" section in their paper.
        \item The paper should point out any strong assumptions and how robust the results are to violations of these assumptions (e.g., independence assumptions, noiseless settings, model well-specification, asymptotic approximations only holding locally). The authors should reflect on how these assumptions might be violated in practice and what the implications would be.
        \item The authors should reflect on the scope of the claims made, e.g., if the approach was only tested on a few datasets or with a few runs. In general, empirical results often depend on implicit assumptions, which should be articulated.
        \item The authors should reflect on the factors that influence the performance of the approach. For example, a facial recognition algorithm may perform poorly when image resolution is low or images are taken in low lighting. Or a speech-to-text system might not be used reliably to provide closed captions for online lectures because it fails to handle technical jargon.
        \item The authors should discuss the computational efficiency of the proposed algorithms and how they scale with dataset size.
        \item If applicable, the authors should discuss possible limitations of their approach to address problems of privacy and fairness.
        \item While the authors might fear that complete honesty about limitations might be used by reviewers as grounds for rejection, a worse outcome might be that reviewers discover limitations that aren't acknowledged in the paper. The authors should use their best judgment and recognize that individual actions in favor of transparency play an important role in developing norms that preserve the integrity of the community. Reviewers will be specifically instructed to not penalize honesty concerning limitations.
    \end{itemize}

\item {\bf Theory assumptions and proofs}
    \item[] Question: For each theoretical result, does the paper provide the full set of assumptions and a complete (and correct) proof?
    \item[] Answer: \answerNA{} 
    \item[] Justification: This thesis does not include formal theoretical results and mathematical proofs. It is mainly supported by experimental and simulation data of the computational model.
    \item[] Guidelines:
    \begin{itemize}
        \item The answer NA means that the paper does not include theoretical results. 
        \item All the theorems, formulas, and proofs in the paper should be numbered and cross-referenced.
        \item All assumptions should be clearly stated or referenced in the statement of any theorems.
        \item The proofs can either appear in the main paper or the supplemental material, but if they appear in the supplemental material, the authors are encouraged to provide a short proof sketch to provide intuition. 
        \item Inversely, any informal proof provided in the core of the paper should be complemented by formal proofs provided in appendix or supplemental material.
        \item Theorems and Lemmas that the proof relies upon should be properly referenced. 
    \end{itemize}

    \item {\bf Experimental result reproducibility}
    \item[] Question: Does the paper fully disclose all the information needed to reproduce the main experimental results of the paper to the extent that it affects the main claims and/or conclusions of the paper (regardless of whether the code and data are provided or not)?
    \item[] Answer: \answerYes{}{} 
    \item[] Justification: We introduced all the parameters related to the experiments in the Method overview and Experiments. And the code is available.
    \item[] Guidelines:
    \begin{itemize}
        \item The answer NA means that the paper does not include experiments.
        \item If the paper includes experiments, a No answer to this question will not be perceived well by the reviewers: Making the paper reproducible is important, regardless of whether the code and data are provided or not.
        \item If the contribution is a dataset and/or model, the authors should describe the steps taken to make their results reproducible or verifiable. 
        \item Depending on the contribution, reproducibility can be accomplished in various ways. For example, if the contribution is a novel architecture, describing the architecture fully might suffice, or if the contribution is a specific model and empirical evaluation, it may be necessary to either make it possible for others to replicate the model with the same dataset, or provide access to the model. In general. releasing code and data is often one good way to accomplish this, but reproducibility can also be provided via detailed instructions for how to replicate the results, access to a hosted model (e.g., in the case of a large language model), releasing of a model checkpoint, or other means that are appropriate to the research performed.
        \item While NeurIPS does not require releasing code, the conference does require all submissions to provide some reasonable avenue for reproducibility, which may depend on the nature of the contribution. For example
        \begin{enumerate}
            \item If the contribution is primarily a new algorithm, the paper should make it clear how to reproduce that algorithm.
            \item If the contribution is primarily a new model architecture, the paper should describe the architecture clearly and fully.
            \item If the contribution is a new model (e.g., a large language model), then there should either be a way to access this model for reproducing the results or a way to reproduce the model (e.g., with an open-source dataset or instructions for how to construct the dataset).
            \item We recognize that reproducibility may be tricky in some cases, in which case authors are welcome to describe the particular way they provide for reproducibility. In the case of closed-source models, it may be that access to the model is limited in some way (e.g., to registered users), but it should be possible for other researchers to have some path to reproducing or verifying the results.
        \end{enumerate}
    \end{itemize}

\item {\bf Open access to data and code}
    \item[] Question: Does the paper provide open access to the data and code, with sufficient instructions to faithfully reproduce the main experimental results, as described in supplemental material?
    \item[] Answer: \answerYes{}{} 
    \item[] Justification: We provide code in the github.
    \item[] Guidelines:
    \begin{itemize}
        \item The answer NA means that paper does not include experiments requiring code.
        \item Please see the NeurIPS code and data submission guidelines (\url{https://nips.cc/public/guides/CodeSubmissionPolicy}) for more details.
        \item While we encourage the release of code and data, we understand that this might not be possible, so “No” is an acceptable answer. Papers cannot be rejected simply for not including code, unless this is central to the contribution (e.g., for a new open-source benchmark).
        \item The instructions should contain the exact command and environment needed to run to reproduce the results. See the NeurIPS code and data submission guidelines (\url{https://nips.cc/public/guides/CodeSubmissionPolicy}) for more details.
        \item The authors should provide instructions on data access and preparation, including how to access the raw data, preprocessed data, intermediate data, and generated data, etc.
        \item The authors should provide scripts to reproduce all experimental results for the new proposed method and baselines. If only a subset of experiments are reproducible, they should state which ones are omitted from the script and why.
        \item At submission time, to preserve anonymity, the authors should release anonymized versions (if applicable).
        \item Providing as much information as possible in supplemental material (appended to the paper) is recommended, but including URLs to data and code is permitted.
    \end{itemize}

\item {\bf Experimental setting/details}
    \item[] Question: Does the paper specify all the training and test details (e.g., data splits, hyperparameters, how they were chosen, type of optimizer, etc.) necessary to understand the results?
    \item[] Answer: \answerYes{} 
    \item[] Justification: We introduced the functions of different parameters in the Method overview.
    \item[] Guidelines:
    \begin{itemize}
        \item The answer NA means that the paper does not include experiments.
        \item The experimental setting should be presented in the core of the paper to a level of detail that is necessary to appreciate the results and make sense of them.
        \item The full details can be provided either with the code, in appendix, or as supplemental material.
    \end{itemize}

\item {\bf Experiment statistical significance}
    \item[] Question: Does the paper report error bars suitably and correctly defined or other appropriate information about the statistical significance of the experiments?
    \item[] Answer: \answerYes{} 
    \item[] Justification: We present the results with error bars in Figure $\ref{Figure 2}$ and conduct a parameter sensitivity analysis.
    \item[] Guidelines:
    \begin{itemize}
        \item The answer NA means that the paper does not include experiments.
        \item The authors should answer "Yes" if the results are accompanied by error bars, confidence intervals, or statistical significance tests, at least for the experiments that support the main claims of the paper.
        \item The factors of variability that the error bars are capturing should be clearly stated (for example, train/test split, initialization, random drawing of some parameter, or overall run with given experimental conditions).
        \item The method for calculating the error bars should be explained (closed form formula, call to a library function, bootstrap, etc.)
        \item The assumptions made should be given (e.g., Normally distributed errors).
        \item It should be clear whether the error bar is the standard deviation or the standard error of the mean.
        \item It is OK to report 1-sigma error bars, but one should state it. The authors should preferably report a 2-sigma error bar than state that they have a 96\% CI, if the hypothesis of Normality of errors is not verified.
        \item For asymmetric distributions, the authors should be careful not to show in tables or figures symmetric error bars that would yield results that are out of range (e.g. negative error rates).
        \item If error bars are reported in tables or plots, The authors should explain in the text how they were calculated and reference the corresponding figures or tables in the text.
    \end{itemize}

\item {\bf Experiments compute resources}
    \item[] Question: For each experiment, does the paper provide sufficient information on the computer resources (type of compute workers, memory, time of execution) needed to reproduce the experiments?
    \item[] Answer: \answerYes{} 
    \item[] Justification: We summarize our computational resources in Experiment.
    \item[] Guidelines:
    \begin{itemize}
        \item The answer NA means that the paper does not include experiments.
        \item The paper should indicate the type of compute workers CPU or GPU, internal cluster, or cloud provider, including relevant memory and storage.
        \item The paper should provide the amount of compute required for each of the individual experimental runs as well as estimate the total compute. 
        \item The paper should disclose whether the full research project required more compute than the experiments reported in the paper (e.g., preliminary or failed experiments that didn't make it into the paper). 
    \end{itemize}
    
\item {\bf Code of ethics}
    \item[] Question: Does the research conducted in the paper conform, in every respect, with the NeurIPS Code of Ethics \url{https://neurips.cc/public/EthicsGuidelines}?
    \item[] Answer: \answerYes{} 
    \item[] Justification: We have read the Code of Ethics and make sure to preserve anonymity.
    \item[] Guidelines:
    \begin{itemize}
        \item The answer NA means that the authors have not reviewed the NeurIPS Code of Ethics.
        \item If the authors answer No, they should explain the special circumstances that require a deviation from the Code of Ethics.
        \item The authors should make sure to preserve anonymity (e.g., if there is a special consideration due to laws or regulations in their jurisdiction).
    \end{itemize}

\item {\bf Broader impacts}
    \item[] Question: Does the paper discuss both potential positive societal impacts and negative societal impacts of the work performed?
    \item[] Answer: \answerYes{} 
    \item[] Justification: We have discuss potential societal impacts in Introduction.
    \item[] Guidelines:
    \begin{itemize}
        \item The answer NA means that there is no societal impact of the work performed.
        \item If the authors answer NA or No, they should explain why their work has no societal impact or why the paper does not address societal impact.
        \item Examples of negative societal impacts include potential malicious or unintended uses (e.g., disinformation, generating fake profiles, surveillance), fairness considerations (e.g., deployment of technologies that could make decisions that unfairly impact specific groups), privacy considerations, and security considerations.
        \item The conference expects that many papers will be foundational research and not tied to particular applications, let alone deployments. However, if there is a direct path to any negative applications, the authors should point it out. For example, it is legitimate to point out that an improvement in the quality of generative models could be used to generate deepfakes for disinformation. On the other hand, it is not needed to point out that a generic algorithm for optimizing neural networks could enable people to train models that generate Deepfakes faster.
        \item The authors should consider possible harms that could arise when the technology is being used as intended and functioning correctly, harms that could arise when the technology is being used as intended but gives incorrect results, and harms following from (intentional or unintentional) misuse of the technology.
        \item If there are negative societal impacts, the authors could also discuss possible mitigation strategies (e.g., gated release of models, providing defenses in addition to attacks, mechanisms for monitoring misuse, mechanisms to monitor how a system learns from feedback over time, improving the efficiency and accessibility of ML).
    \end{itemize}
    
\item {\bf Safeguards}
    \item[] Question: Does the paper describe safeguards that have been put in place for responsible release of data or models that have a high risk for misuse (e.g., pretrained language models, image generators, or scraped datasets)?
    \item[] Answer: \answerNA{} 
    \item[] Justification: Our work poses no such risks.
    \item[] Guidelines:
    \begin{itemize}
        \item The answer NA means that the paper poses no such risks.
        \item Released models that have a high risk for misuse or dual-use should be released with necessary safeguards to allow for controlled use of the model, for example by requiring that users adhere to usage guidelines or restrictions to access the model or implementing safety filters. 
        \item Datasets that have been scraped from the Internet could pose safety risks. The authors should describe how they avoided releasing unsafe images.
        \item We recognize that providing effective safeguards is challenging, and many papers do not require this, but we encourage authors to take this into account and make a best faith effort.
    \end{itemize}

\item {\bf Licenses for existing assets}
    \item[] Question: Are the creators or original owners of assets (e.g., code, data, models), used in the paper, properly credited and are the license and terms of use explicitly mentioned and properly respected?
    \item[] Answer: \answerNA{} 
    \item[] Justification: We develop the model from scratch.
    \item[] Guidelines:
    \begin{itemize}
        \item The answer NA means that the paper does not use existing assets.
        \item The authors should cite the original paper that produced the code package or dataset.
        \item The authors should state which version of the asset is used and, if possible, include a URL.
        \item The name of the license (e.g., CC-BY 4.0) should be included for each asset.
        \item For scraped data from a particular source (e.g., website), the copyright and terms of service of that source should be provided.
        \item If assets are released, the license, copyright information, and terms of use in the package should be provided. For popular datasets, \url{paperswithcode.com/datasets} has curated licenses for some datasets. Their licensing guide can help determine the license of a dataset.
        \item For existing datasets that are re-packaged, both the original license and the license of the derived asset (if it has changed) should be provided.
        \item If this information is not available online, the authors are encouraged to reach out to the asset's creators.
    \end{itemize}

\item {\bf New assets}
    \item[] Question: Are new assets introduced in the paper well documented and is the documentation provided alongside the assets?
    \item[] Answer: \answerYes{} 
    \item[] Justification: Our paper does not release new assets.
    \item[] Guidelines:
    \begin{itemize}
        \item The answer NA means that the paper does not release new assets.
        \item Researchers should communicate the details of the dataset/code/model as part of their submissions via structured templates. This includes details about training, license, limitations, etc. 
        \item The paper should discuss whether and how consent was obtained from people whose asset is used.
        \item At submission time, remember to anonymize your assets (if applicable). You can either create an anonymized URL or include an anonymized zip file.
    \end{itemize}

\item {\bf Crowdsourcing and research with human subjects}
    \item[] Question: For crowdsourcing experiments and research with human subjects, does the paper include the full text of instructions given to participants and screenshots, if applicable, as well as details about compensation (if any)? 
    \item[] Answer: \answerNA{} 
    \item[] Justification: Our paper does not involve crowdsourcing nor research with human subjects.
    \item[] Guidelines:
    \begin{itemize}
        \item The answer NA means that the paper does not involve crowdsourcing nor research with human subjects.
        \item Including this information in the supplemental material is fine, but if the main contribution of the paper involves human subjects, then as much detail as possible should be included in the main paper. 
        \item According to the NeurIPS Code of Ethics, workers involved in data collection, curation, or other labor should be paid at least the minimum wage in the country of the data collector. 
    \end{itemize}

\item {\bf Institutional review board (IRB) approvals or equivalent for research with human subjects}
    \item[] Question: Does the paper describe potential risks incurred by study participants, whether such risks were disclosed to the subjects, and whether Institutional Review Board (IRB) approvals (or an equivalent approval/review based on the requirements of your country or institution) were obtained?
    \item[] Answer: \answerNA{} 
    \item[] Justification: Our paper does not involve crowdsourcing nor research with human subjects.
    \item[] Guidelines:
    \begin{itemize}
        \item The answer NA means that the paper does not involve crowdsourcing nor research with human subjects.
        \item Depending on the country in which research is conducted, IRB approval (or equivalent) may be required for any human subjects research. If you obtained IRB approval, you should clearly state this in the paper. 
        \item We recognize that the procedures for this may vary significantly between institutions and locations, and we expect authors to adhere to the NeurIPS Code of Ethics and the guidelines for their institution. 
        \item For initial submissions, do not include any information that would break anonymity (if applicable), such as the institution conducting the review.
    \end{itemize}

\item {\bf Declaration of LLM usage}
    \item[] Question: Does the paper describe the usage of LLMs if it is an important, original, or non-standard component of the core methods in this research? Note that if the LLM is used only for writing, editing, or formatting purposes and does not impact the core methodology, scientific rigorousness, or originality of the research, declaration is not required.
    \item[] Answer: \answerNA{} 
    \item[] Justification: The core method development in this research does not involve LLMs as any important, original, or non-standard components.
    \item[] Guidelines:
    \begin{itemize}
        \item The answer NA means that the core method development in this research does not involve LLMs as any important, original, or non-standard components.
        \item Please refer to our LLM policy (\url{https://neurips.cc/Conferences/2025/LLM}) for what should or should not be described.
    \end{itemize}

\end{enumerate}


\newpage

\appendix

\section{Discrimination Accuracy over Training Epochs under Different Noise Intensities}

\begin{figure}[htbp] 
    \centering 
    \includegraphics[width=1\textwidth]{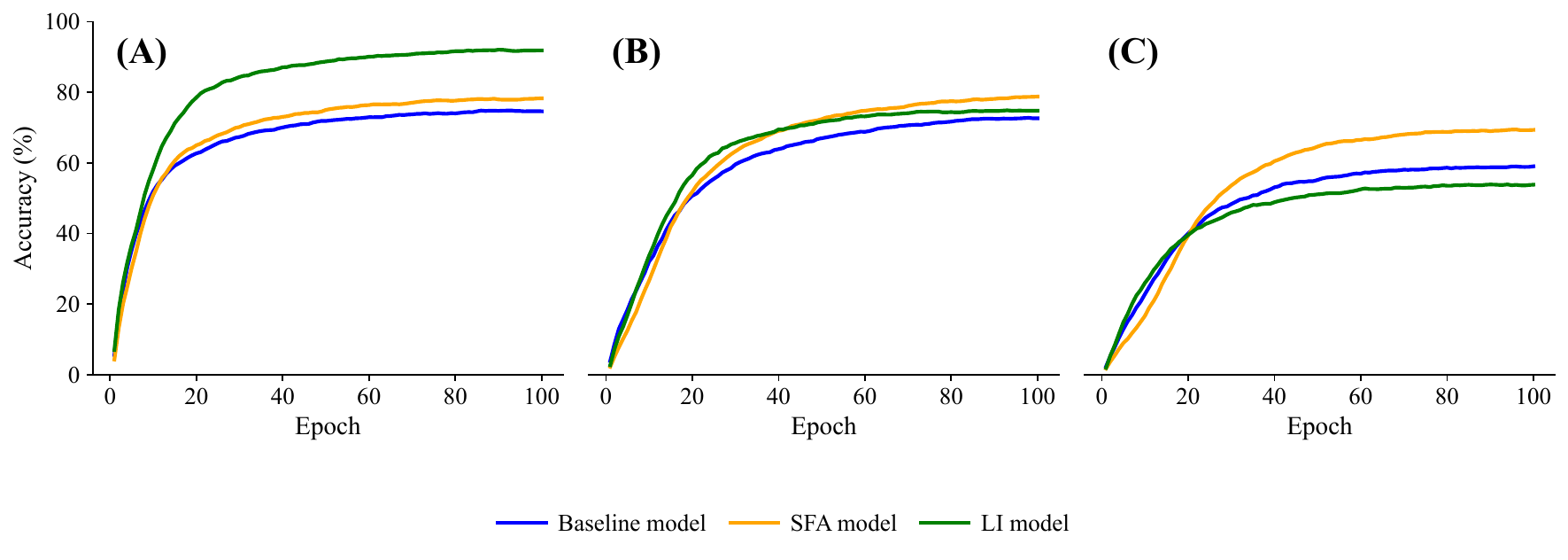} 

    \caption{\textbf{Evolution of discrimination accuracy over training epochs under different noise intensities.} The subfigures show results for: (A) N.I. $= 0.1$, (B) N.I. $= 0.2$, and (C) N.I. $= 0.3$.}
    \label{fig6} 
\end{figure}

The main text primarily presents the final discrimination accuracy values under specific experimental conditions. To better illustrate the learning dynamics, this section includes results showing the time course of recognition accuracy for the three models (Baseline, LI, and SFA models) as they learn to discriminate 1,000 odor classes under varying noise intensities (N.I. $= 0.1$, $0.2$, and $0.3$), as shown in Figure $\ref{fig6}$.

\section{Model Performance with Enhanced Strength of Input Signal}
\label{Performance}
The experiments described in the main text imposed certain limits on the strength of the input odor signals to better isolate performance differences attributable to the LI and SFA mechanisms. Under conditions with a large number of odor categories and substantial noise interference, this constraint led to relatively low absolute accuracy values across all models. In this section, we demonstrate that a moderate increase in input signal strength can significantly enhance overall discrimination performance.

\begin{table}[htbp]
\centering
\caption{\textbf{Comparative performance of the Baseline, SFA, and LI models for 10,000-class odor discrimination under different noise intensities with enhanced strength of input signal.}}
\begin{tabular}{lc|ccc}
\hline
 & N.I. & Baseline & LI & SFA \\
\hline
\multirow{7}{*}{\begin{tabular}[c]{@{}l@{}}Acc.\\ (\%)\end{tabular}}

& 0 & 99.82 & 99.99 & 99.97 \\
& 0.05 & 98.46 & 99.90 & 99.45 \\
& 0.1  & 96.87 &  98.59& 98.14 \\
& 0.15 & 90.13 & 86.58 & 93.27 \\
& 0.2  & 69.80 & 60.79 & 81.31 \\
& 0.25 & 45.21 & 34.17 & 60.07 \\
& 0.3  & 25.94 & 18.40 & 37.45 \\
\hline
\end{tabular}
\label{tab2}
\end{table}

Table 3 summarizes the final discrimination accuracy of the Baseline, LI, and SFA models after increasing the input signal strength. Enhancing the input signal strength led to substantial accuracy improvements for all models across the tested conditions. For example, with 10,000 odor classes and noise $= 0.1$, the accuracy of the Baseline model rose from $21.80\%$ to $96.87\%$.

\end{document}